\documentclass{article}

\usepackage{PRIMEarxiv}

\usepackage[utf8]{inputenc} 
\usepackage[T1]{fontenc}    
\usepackage{url}            
\usepackage{booktabs}       
\usepackage{amsfonts}       
\usepackage{nicefrac}       
\usepackage{microtype}      
\usepackage{lipsum}
\usepackage{fancyhdr}       
\usepackage{graphicx}       
\usepackage{amsmath}
\usepackage{multirow}

\graphicspath{{media/}}     

\usepackage{bm}
\usepackage{adjustbox}

\usepackage{algorithm}
\usepackage{algorithmic}
\usepackage{amsthm}
\usepackage[position=top]{subfig}

\newtheorem{defn}{Definition}[section]
\newtheorem{prop}{Proposition}[section]
\usepackage{newfloat}
\usepackage{listings}
\floatstyle{ruled}
\newfloat{listing}{tb}{lst}{}
\floatname{listing}{Listing}
\usepackage{bibentry}
\usepackage{xcolor}

\title{CompoundE: Knowledge Graph Embedding with Translation, Rotation and Scaling Compound Operations}

\author{
  Xiou Ge \\
  University of Southern California \\
  Los Angeles, USA\\
  \texttt{xiouge@usc.edu} \\
   \And
  Yun-Cheng Wang \\
  University of Southern California \\
  Los Angeles, USA\\
  \texttt{yunchenw@usc.edu} \\
   \And
  Bin Wang \\
  National University of Singapore \\
  Singapore\\
  \texttt{bwang28c@gmail.com} \\
   \And
  C.-C. Jay Kuo \\
  University of Southern California \\
  Los Angeles, USA\\
  \texttt{cckuo@sipi.usc.edu} \\
}

\begin{document}
\maketitle

\begin{abstract}
Translation, rotation, and scaling are three commonly used geometric
manipulation operations in image processing. Besides, some of them are
successfully used in developing effective knowledge graph embedding
(KGE) models such as TransE and RotatE.  Inspired by the synergy, we
propose a new KGE model by leveraging all three operations in this work.
Since translation, rotation, and scaling operations are cascaded to form
a compound one, the new model is named CompoundE. By casting CompoundE
in the framework of group theory, we show that quite a few
scoring-function-based KGE models are special cases of CompoundE.
CompoundE extends the simple distance-based relation to
relation-dependent compound operations on head and/or tail entities.  To
demonstrate the effectiveness of CompoundE, we conduct experiments on
three popular KG completion datasets. Experimental
results show that CompoundE consistently achieves the state-of-the-art
performance. 
\end{abstract}

\keywords{Knowledge graph embedding \and Compounding Operations}

\renewcommand{\figurename}{Figure}
\renewcommand{\tablename}{Table}

\section{Introduction}\label{sec:introduction}

Knowledge graphs (KGs) such as DBpedia \cite{auer2007dbpedia}, YAGO
\cite{suchanek2007yago}, NELL \cite{carlson2010toward}, Wikidata
\cite{vrandevcic2014wikidata}, Freebase \cite{bollacker2008freebase},
and ConceptNet \cite{speer2017conceptnet} have been created and made
available to the public to facilitate research on KG modeling and
applications.  KG representation learning, also known as knowledge graph
embedding (KGE), has been intensively studied in recent years.  Yet, it
remains to be one of the most fundamental problems in Artificial
Intelligence (AI) and Data Engineering research.  KGE is critical to
many downstream applications such as multihop reasoning
\cite{shen2020modeling, du2021cogkr}, KG alignment
\cite{chen2016multilingual, largeEA}, entity classification \cite{ge2022core}, etc. 

Triples, denoted by $(h,r,t)$, are basic elements of a KG, where $h$ and
$t$ are head and tail entities while $r$ is the relation connecting
them.  For instance, the fact ``Los Angeles is located in the USA'' can
be encoded as a triple (Los Angeles, \textbf{isLocatedIn}, USA).  The
link prediction (or KG completion) task is often used to evaluate the
effectiveness of KGE models. That is, the task is to predict $t$ with
given $h$ and $r$, or to predict $h$ with given $r$ and $t$. KGE models
are evaluated based on how well the prediction matches the ground truth. 

There are several challenges in the design of good KGE models.  First,
real-world KGs often contain a large number of entities. It is
impractical to have high-dimensional embeddings due to device memory
constraints. Yet, the performance of KGE models may deteriorate
significantly in low-dimensional settings. How to design a KGE model
that is effective in low-dimensional settings is not trivial. Second,
complex relation types (e.g., hierarchical relations, surjective
relations, antisymmetric relations, etc.) remain difficult to model.
The link prediction performance for 1-N, N-1, and N-N relations is challenging for many existing KGE models.  The relation ``isLocatedIn'' is an example
of a N-1 relation. Since there are many cities other than ``Los
Angeles'' also located in the USA, it is not easy to encode these
relations effectively.  Third, each of extant KGE models has its own
strengths and weaknesses. It is desired yet unclear how to design a KGE
model that leverages strengths of some models and complements weaknesses
of others. 

Geometric manipulation operations such as translation and rotation have
been used to build effective knowledge graph embedding (KGE) models
(e.g., TransE, RotatE).  Inspired by their success, we look for
generalized geometric manipulations in image processing
\cite{pratt2013introduction}.  To this end, translation, rotation, and
scaling are three common geometric manipulation operations. Furthermore,
they can be cascaded to yield a generic compound operation that finds
numerous applications.  Examples include image warping
\cite{wolberg1990digital}, image morphing \cite{seitz1996view}, and
robot motion planning \cite{lavalle2006planning}. Motivated by the
synergy, we propose a new KGE model to address the above-mentioned
challenges. Since translation, rotation, and scaling operations are
cascaded to form a compound operation, the proposed KGE model is named
CompoundE. Compound operations inherit many desirable properties from
the affine group, allowing CompoundE to model complex relations in
different KGs.  Moreover, since geometric transformations can be
composed in any order, CompoundE has a large number of design variations
to choose from. One can select the optimal CompoundE variant that best
suits the characteristics of an individual dataset. 

There are four main contributions of this work. They are summarized below.
\begin{itemize}
\item We present a novel KG embedding model called CompoundE, which
combines three fundamental operations in the affine group and offers a
wide range of designs. 
\item It is proved mathematically that CompoundE can handle complex
relation types in KG thanks to unique properties of the affine group. 
\item We conduct extensive experiments on three popular KG completion
datasets; namely FB15k-237, WN18RR, and ogbl-wikikg2.  Experimental
results show that CompoundE achieves state-of-the-art performance.
\item Against large-scale datasets containing millions of entities under
the memory constraint, CompoundE outperforms other benchmarking methods
by a big margin with fewer parameters. 
\end{itemize}

The rest of this paper is organized as follows.  Recent KGE models for
both distance-based and semantic matching-based categories are first
reviewed in Section \ref{sec:related_work}. Then, we present CompoundE,
show its relationship with previous KG embedding models, and explain the
reason why it can model complex relations well in Section
\ref{sec:method}. Versatile transformations introduced by CompoundE lead
to better performance. To demonstrate its effectiveness, we conduct
extensive experiments on the link prediction task in Section
\ref{sec:experiments}.  Finally, concluding remarks are given and
possible extensions are suggested in Section \ref{sec:conclusion}. 

\begin{figure*}[t]
\centering
\includegraphics[width=\textwidth]{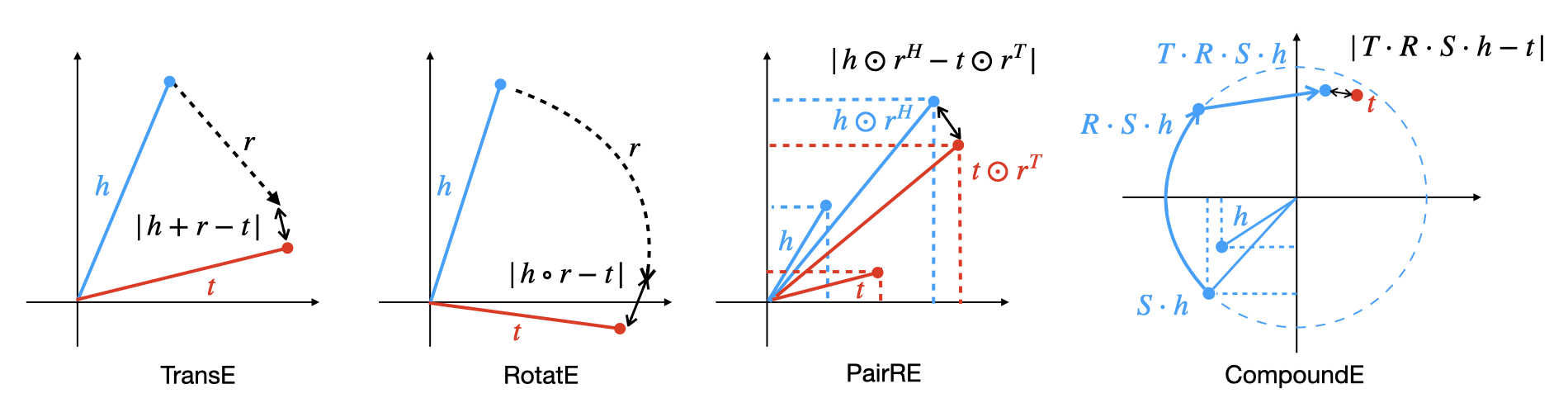}
\caption{An illustration of previous distance-based KGE models 
and CompoundE.}\label{overall_framework}
\end{figure*}

\section{Related Work}\label{sec:related_work}

\subsection{Distance-based Models}

The distance-based strategy offers a popular methodology in developing
KGE models. The main idea is to model a relation as a transformation
that places head entity vectors in the proximity of their corresponding
tail entity vectors, and vice versa. For a given triple, $(h,r,t)$, the
goal is to minimize the distance between $h$ and $t$ vectors after the
transformation introduced by $r$.  TransE \cite{bordes2013translating}
is one of the earlier KGE models that interpret relations between
entities as translation operations in the vector space. Drawbacks of
TransE include inability to model one-to-many, many-to-one,
many-to-many, and symmetric relations.  RotatE \cite{sun2018rotate}
models entities in the complex vector space and interprets a relation as
a rotation instead of a translation. The self-adversarial negative
sampling contributes to RotatE's performance improvement as compared to
its predecessors.  Instead of rotating heads to match tails, PairRE
\cite{chao-etal-2021-pairre} performs transformations on both heads and
tails. Specifically, head and tail entities had a Hadamard product with
their respective weight vectors $r^H$ and $r^T$. This elementwise
multiplication is nothing but the scaling operation. LinearRE
\cite{peng2020lineare} offers a similar model but adds a translation
component between the scaled head and tail entities.  TorusE
\cite{ebisu2019generalized} projects embedding in an $n$-dimensional
torus space and optimizes the generalized translation-based embedding.
MuRP \cite{balazevic2019multi} and ROTH \cite{chami2020low} leverage
the power of hyperbolic geometry to better model the hierarchical
structure in KGs. 

\subsection{Semantic-Matching-based Models}

Another related idea of developing KGE models is to measure the semantic
matching score. RESCAL \cite{nickel2011three} adopts a bilinear scoring
function as the objective in solving a three-way rank-$r$ matrix
factorization problem. DistMult \cite{yang2014embedding} also uses a
bilinear scoring function but enforces the relation embedding matrix to
be diagonal. ANALOGY \cite{liu2017analogical} demands the relation
embedding matrix to be orthonormal and commutative by imposing
regularization constraints. ComplEx \cite{trouillon2016complex} extends
the bilinear product score to the complex vector space so as to model
antisymmetric relations more effectively. SimplE \cite{kazemi2018simple} and TuckER
\cite{balavzevic2019tucker} further model 3D tensor factorization with Canonical Polyadic decomposition and Tucker decomposition, respectively. QuatE
\cite{zhang2019quaternion} and DualE \cite{cao2021dual} extends from
the complex representation to the hypercomplex representation with 4
degrees of freedom to gain more expressive rotational capability. 

\subsection{Classifier-based Models}

Another popular approach to build KGE models is based on the
classification framework.  For example, a Multilayer Perceptron (MLP)
\cite{dong2014knowledge} is used to measure the likelihood of unseen
triples for link prediction.  NTN \cite{socher2013reasoning} adopts a
bilinear tensor neural layer to model interactions between entities and
relations of triples. ConvE \cite{dettmers2018convolutional} reshapes
and stacks the head entity and the relation vector to form a 2D shape
data, applies Convolutional Neural Networks (CNNs) to extract features,
and uses extracted features to interact with tail embedding. R-GCN
\cite{schlichtkrull2018modeling} applies a Graph Convolutional Network
(GCN) and considers the neighborhood of each entity equally. CompGCN
\cite{vashishth2020compositionbased} performs a composition operation
over each edge in the neighborhood of a central node. The composed
embeddings are then convolved with specific filters representing the
original and the inverse relations, respectively. M-DCN
\cite{zhang2020multi} adopts a multi-scale dynamic convolutional
network to model complex relations such as 1-N, N-1, and N-N relations.
All the above mentioned examples are based on neural networks. Yet, there
is a non-neural-network classification method, called KGBoost
\cite{wang2022kgboost}, developed recently. KGBoost proposes a novel
negative sampling method and uses the XGBoost classifier for link
prediction. 

\subsection{Entity-Transformation-based Models}

Another line of work attempts to improve baseline models such as TransE
by adding a relation-specific transformation. TransH
\cite{wang2014knowledge} enables the relation-specific entity
representation by projecting each entity to a relation-specific
hyperplane. TransR \cite{lin2015learning} models relations and entities
in two different spaces. TransD \cite{ji2015knowledge} comes up with
dynamic mapping matrices using relation and entity projection vectors.
TranSparse \cite{ji2016knowledge} enforces the relation projection
matrix to be sparse. SFBR \cite{liang2021semantic} introduces a
semantic filter that includes a scaling and shift component. ReflectE
\cite{zhang2022knowledge} models a relation as the Householder
reflection. 

\section{Method}\label{sec:method}

Translation, rotation, and scaling transformations appear frequently in
engineering applications. In image processing, a cascade of translation,
rotation, and scaling operations offers a set of image manipulation
techniques. Such compound operations can be used to develop a new KGE
model called CompoundE. We define CompoundE in Section
\ref{subsec:definition}, explain that it belongs to the affine group in
Section \ref{subsec:affine}, and show that it is a generalized form of
TransE, RotatE, PairRE, and a few other distance-based embedding models
in Section \ref{subsec:connection}. Furthermore, we discuss the
properties of CompoundE in Section \ref{subsec:properties}. 

\subsection{Definition of CompoundE}\label{subsec:definition}
Three forms of CompoundE scoring function can be written as
\begin{itemize}
    \item CompoundE-Head
\begin{equation}
    f_r(h,t) = \|\mathbf{T_r\cdot R_r\cdot S_r\cdot h - t }\|,
\end{equation}
\item CompoundE-Tail
\begin{equation}
    f_r(h,t) = \|\mathbf{h - \hat{T}_r\cdot \hat{R}_r\cdot \hat{S}_r\cdot t }\|,
\end{equation}
\item CompoundE-Full
\begin{equation}
    f_r(h,t) = \|\mathbf{T_r\cdot R_r\cdot S_r\cdot h - \hat{T}_r\cdot \hat{R}_r\cdot \hat{S}_r\cdot t }\|,
\end{equation}
\end{itemize}
where $\mathbf{h, t}$ denote head and tail entity embeddings,
$\mathbf{T_r, R_r, S_r}$ denote the translation, rotation, and scaling
operations for the head entity embedding, and $\mathbf{\hat{T}_r,
\hat{R}_r, \hat{S}_r}$ denote the counterparts for the tail entity
embedding, respectively.  These constituent operators are
relation-specific.  To generalize, any order or subset of translation,
rotation, and scaling component can be a valid instance of CompoundE.
Since matrix multiplications are non-commutative, different orders of
cascading the constituent operators result in distinct CompoundE
operators.  
We illustrate different ways
of cascading geometric transformations to compose distinct CompoundE
operators in Fig. \ref{compoundE_permute} of the appendix. 

We follow RotatE's negative sampling loss and the self-adversarial training
strategy. The loss function of CompoundE can be written as
\begin{equation}
     L_{\textnormal{KGE}}=-\log\sigma(\zeta_1-f_r(h, t))-\sum_{i=1}^np(h'_i,r,t'_i)\log\sigma(f_r(h'_i,t'_i)-\zeta_1),
\end{equation}
where $\sigma$ is the sigmoid function, $\zeta_1$ is a fixed margin
hyperparameter, $(h'_i,r,t'_i)$ is the $i$th negative triple, and
$p(h'_i,r,t'_i)$ is the probability of drawing negative triple
$(h'_i,r,t'_i)$. Given a positive triple, $(h_i,r,t_i)$,
the negative sampling distribution is
\begin{equation}
p(h'_j,r,t'_j|\{(h_i,r,t_i)\})=\frac{\exp \alpha_1 f_r(h'_j, t'_j)}
{\sum_i \exp \alpha_1 f_r(h'_i, t'_i)},
\end{equation}
where $\alpha_1$ is the temperature of sampling.

\subsection{CompoundE As An Affine Group}\label{subsec:affine}

Most analysis in previous work \cite{cao2022geometry} was restricted to the
Special Euclidean Group $\mathbf{SE}(n)$. Yet, we will show that
CompoundE is not a special Euclidean group but an affine group.  To
proceed, we need to introduce the lie group and three special groups
formally below. 
\begin{defn}
A Lie group is a continuous group that is also a differentiable manifold.
\end{defn}
Several Lie group examples are given below.
\begin{itemize}
\item The real vector space, $\mathbb{R}^n$, with the canonical addition
as the group operation. 
\item The real vector space excluding zero,
$(\mathbb{R}\backslash\{0\})$, with the element-wise multiplication as
the group operation. 
\item The general linear group, $GL_n(\mathbb{R})$, with the the
canonical matrix multiplication as the group operation. 
\end{itemize}
Furthermore, the following three special groups are commonly used.
\begin{defn}
The special orthogonal group is defined as
\end{defn}
\begin{equation}
\mathbf{SO}(n) = \left \{ \mathbf{A}\in \mathbf{GL}_n(\mathbb{R}) \bigg|
\mathbf{A}^T\mathbf{A} = \mathbf{I} \wedge \det \mathbf{A}=1 \right \}.
\end{equation}
\begin{defn}
The special Euclidean group is defined as
\end{defn}
\begin{equation}\label{eq:se}
\mathbf{SE}(n) = \left \{ \mathbf{A} \bigg| \mathbf{A}=
    \begin{bmatrix}
      \mathbf{R} & \mathbf{v} \\
      \mathbf{0} & 1
    \end{bmatrix}, \mathbf{R}\in \mathbf{SO}(n), \mathbf{v}\in \mathbb{R}^n\right \}.
\end{equation}
\begin{defn}
The affine group is defined as
\end{defn}
\begin{equation}\label{eq:aff}
    \mathbf{Aff}(n) = \left \{ \mathbf{M} \bigg| \mathbf{M}=
    \begin{bmatrix}
      \mathbf{A} & \mathbf{v} \\
      \mathbf{0} & 1
    \end{bmatrix}, \mathbf{A}\in \mathbf{GL}_n(\mathbb{R}), \mathbf{v}\in 
    \mathbb{R}^n\right \}.
\end{equation}
By comparing Eqs. (\ref{eq:se}) and (\ref{eq:aff}), we see that
$\mathbf{SE}(n)$ is a subset of $\mathbf{Aff}(n)$. 

Without loss of generality, consider $n=2$. If $\mathbf{M} \in \mathbf{Aff}(2)$, we have
\begin{equation}
    \mathbf{M} = 
  \begin{bmatrix}
    \mathbf{A} & \mathbf{v} \\
    \mathbf{0} & 1
  \end{bmatrix} , \mathbf{A}\in\mathbb{R}^{2\times2} , \mathbf{v}\in \mathbb{R}^{2}.
\end{equation}
The 2D translational matrix can be written as
\begin{equation}
    \mathbf{T} = \begin{bmatrix}
    1 & 0 & v_x \\
    0 & 1 & v_y \\
    0 & 0 & 1
  \end{bmatrix},
\end{equation}
while the 2D rotational matrix can be expressed as
\begin{equation}
    \mathbf{R} = \begin{bmatrix}
    \cos(\theta) & -\sin(\theta) & 0 \\
    \sin(\theta) & \cos(\theta) & 0 \\
    0 & 0 & 1
  \end{bmatrix}.
\end{equation}
It is easy to verify that they are both special Euclidean groups (i.e.
$\mathbf{T}\in \mathbf{SE}(2)$ and $\mathbf{R}\in \mathbf{SE}(2)$).  On the
other hand, the 2D scaling matrix is in form of
\begin{equation}
    \mathbf{S} = \begin{bmatrix}
    s_x & 0 & 0 \\
    0 & s_y & 0 \\
    0 & 0 & 1
  \end{bmatrix}.
\end{equation}
It is not a special Euclidean group but an affine group of $n=2$
(i.e., $\mathbf{S}\in\mathbf{Aff}(2)$).

Compounding translation and rotation operations, we can get a
transformation in the special Euclidean group,
\begin{equation}
\begin{aligned}
    \mathbf{T}\cdot \mathbf{R} &= \begin{bmatrix}
    1 & 0 & v_x \\
    0 & 1 & v_y \\
    0 & 0 & 1
  \end{bmatrix}\begin{bmatrix}
    \cos(\theta) & -\sin(\theta) & 0 \\
    \sin(\theta) & \cos(\theta) & 0 \\
    0 & 0 & 1
  \end{bmatrix} \\&= \begin{bmatrix}
    \cos(\theta) & -\sin(\theta) & v_x \\
    \sin(\theta) & \cos(\theta) & v_y \\
    0 & 0 & 1
  \end{bmatrix} \in \mathbf{SE}(2).
\end{aligned}
\end{equation}
Yet, if we add the scaling operation, the compound will belong to the
Affine group. One of such compound operator can be written as
\begin{equation}
\begin{aligned}
    \mathbf{T}\cdot \mathbf{R}\cdot \mathbf{S} &= \begin{bmatrix}
    1 & 0 & t_x \\
    0 & 1 & t_y \\
    0 & 0 & 1
  \end{bmatrix}\begin{bmatrix}
    \cos(\theta) & -\sin(\theta) & 0 \\
    \sin(\theta) & \cos(\theta) & 0 \\
    0 & 0 & 1
  \end{bmatrix}\begin{bmatrix}
    s_x & 0 & 0 \\
    0 & s_y & 0 \\
    0 & 0 & 1
  \end{bmatrix} \\&= \begin{bmatrix}
    s_x\cos(\theta) & -s_y\sin(\theta) & v_x \\
    s_x\sin(\theta) & s_y\cos(\theta) & v_y \\
    0 & 0 & 1
  \end{bmatrix} \in \mathbf{Aff}(2).
\end{aligned}
\end{equation}
When $s_x\neq0$ and $s_y\neq0$, the compound operator is invertible.
It can be written in form of
\begin{equation}
  \mathbf{M}^{-1} = 
  \begin{bmatrix}
    \mathbf{A}^{-1} & -\mathbf{A}^{-1}\mathbf{v} \\
    \mathbf{0} & 1
  \end{bmatrix}.
\end{equation}

\subsection{Relation with Other Distance-based KGE Models}\label{subsec:connection}

CompoundE is actually a general form of quite a few distance-based KGE
models. That is, we can derive their scoring functions from that of
CompoundE by setting translation, scaling, and rotation operations to
certain forms. Four examples are given below. \\
\noindent
{\bf Derivation of TransE.} We begin with CompoundE-Head and set its 
rotation component to identity matrix $\mathbf{I}$ and scaling parameters
to $\mathbf{1}$. Then, we get the scoring function of TransE as
\begin{equation}
f_r(h,t) = \|\mathbf{T_r\cdot I \cdot \text{diag}(1)\cdot h - t }\| = 
\|\mathbf{ h +r - t }\|.
\end{equation}

\noindent
{\bf Derivation of RotatE.} We can derive the scoring function of
RotatE from CompoundE-Head by setting the translation component to
$\mathbf{I}$ (translation vector $\mathbf{t = 0}$) and scaling component
to $\mathbf{1}$. 
\begin{equation}
f_r(h,t) = \|\mathbf{I\cdot R_r\cdot \text{diag}(1)\cdot h - t }\| = 
\|\mathbf{ h \circ r - t }\|.
\end{equation}
\noindent
{\bf Derivation of PairRE.} CompoundE-Full can be reduced to PairRE by
setting both translation and rotation component to $\mathbf{I}$, for
both head and tail transformation. 
\begin{equation}
f_r(h,t) = \|\mathbf{I\cdot I\cdot S_r\cdot h - I\cdot I\cdot \hat{S}_r\cdot t }\| 
= \|\mathbf{h\odot r^H - t\odot r^T}\|.
\end{equation}
\noindent
{\bf Derivation of LinearRE.} We can add back the translation component
for the head transformation:
\begin{equation}
\begin{aligned}
f_r(h,t) &= \|\mathbf{T_r\cdot I\cdot S_r\cdot h - I\cdot I\cdot \hat{S}_r\cdot t }\| = \|\mathbf{h\odot r^H +r- t\odot r^T}\|.
\end{aligned}
\end{equation}

\subsection{Properties of CompoundE}\label{subsec:properties}

CompoundE has a richer set of operations and, therefore, it is more
powerful than previous KGE models in modeling complex relations such as
1-to-N, N-to-1, and N-to-N relations in KG datasets. Modeling these
relations are important since more than 98$\%$ of triples in FB15k-237
and WN18RR datasets involves complex relations.  The importance of
complex relations modeling is illustrated by two examples below. First, there is a need to distinguish different outcomes of relation compositions when modeling non-commutative relations.
That is $r_1 \cdot r_2
\rightarrow r_3$ while $r_2 \cdot r_1 \rightarrow r_4$.  For instance,
$r_1$, $r_2$, $r_3$ and $r_4$ denote $\textbf{isFatherOf}$,
$\textbf{isMotherOf}$, $\textbf{isGrandfatherOf}$ and
$\textbf{isGrandmotherOf}$, respectively.  TransE and RotatE cannot make
such distinction since they are based on commutative relation
embeddings. Second, to capture the hierarchical structure of relations,
it is essential to build a good model for sub-relations. For example,
$r_1$ and $r_2$ denote $\textbf{isCapitalCityOf}$ and
$\textbf{cityLocatedInCountry}$, respectively. Logically,
$\textbf{isCapitalCityOf}$ is a sub-relation of
$\textbf{cityLocatedInCountry}$ because if $(h, r_1, t)$ is true, then
$(h, r_2, t)$ must be true.  We will prove that CompoundE is capable of
modeling symmetric/antisymmetric, inversion, transitive,
commutative/non-commutative, and sub-relations in Section 6.3 of the appendix. 

\begin{table}[h]
 \caption{Filtered ranking of link prediction on ogbl-wikikg2.}
  \begin{center}
    \label{tab:wikikg2_overall}
    \begin{tabular}{c|c|c|c} 
      \hline
      \textbf{Datasets} & \multicolumn{3}{c}{\textbf{ogbl-wikikg2}} \\
      \hline
      \multirow{2}{*}{\textbf{Metrics}} & \multirow{2}{*}{\textbf{Dim}} & \textbf{Valid} & \textbf{Test}\\
      & & \textbf{MRR} & \textbf{MRR} \\
      \hline
      AutoSF+NodePiece & - & 0.5806 & 0.5703 \\
      ComplEx-RP & 50 & \underline{0.6561} & \underline{0.6392} \\
      \hline
      TransE & 500 & 0.4272 & 0.4256 \\
      DistMult & 500 & 0.3506 & 0.3729 \\
      ComplEx & 250 & 0.3759 & 0.4027 \\
      RotatE & 250 & 0.4353 & 0.4353 \\
      PairRE & 200 & 0.5423 & 0.5208 \\
      TripleRE & 200 & 0.6045 & 0.5794\\ \hline
      CompoundE & 100 & \textbf{0.6704} & \textbf{0.6515} \\ \hline
    \end{tabular}
  \end{center}
\end{table}

\begin{table*}[ht!]
  \caption{Filtered ranking of link prediction for FB15k-237 and WN18RR.}
  \begin{center}
    \label{tab:fb15k-237_wn18rr_overall}
    \begin{tabular}{c|cccc|cccc} 
      \hline
      \textbf{Datasets} & \multicolumn{4}{c|}{\textbf{FB15K-237}} & \multicolumn{4}{c}{\textbf{WN18RR}}\\
      \hline
      \textbf{Metrics} & \textbf{MRR} & \textbf{Hit@1} & \textbf{Hit@3} & \textbf{Hit@10} & \textbf{MRR} & \textbf{Hit@1} & \textbf{Hit@3} & \textbf{Hit@10}\\
      \hline
      TransE \cite{bordes2013translating} & 0.294 & - & - & 0.465 & 0.226 & - & - & 0.501 \\
      DistMult \cite{yang2014embedding} & 0.241 & 0.155 & 0.263 & 0.419 & 0.430 & 0.390 & 0.440 & 0.490 \\
      ComplEx \cite{trouillon2016complex} & 0.247 & 0.158 & 0.275 & 0.428 & 0.440 & 0.410 & 0.460 & 0.510 \\
      RotatE \cite{sun2018rotate} & 0.338 & 0.241 & 0.375 & 0.533 & 0.476 & 0.428 & 0.492 & 0.571 \\
      TorusE \cite{Ebisu2020GeneralizedTE} & 0.316 & 0.217 & 0.335 & 0.484 & 0.453 & 0.422 & 0.464 & 0.512 \\
      TuckER \cite{balavzevic2019tucker} & 0.358 & 0.266 & 0.394 & 0.544 & 0.470 & 0.443 & 0.482 & 0.526 \\
      AutoSF \cite{zhang2020autosf} & 0.360 & 0.267 & - & \underline{0.552} & 0.490 & 0.451 & - & 0.567 \\
      RotatE3D \cite{gao2020rotate3d} & 0.347 & 0.250 & 0.385 & 0.543 & 0.489 & 0.442 & \underline{0.505} & \textbf{0.579} \\
      MQuadE \cite{yu2021mquade} & 0.356 & 0.260 & 0.392 & 0.549 & - & - & - & - \\
      PairRE \cite{chao-etal-2021-pairre} & 0.351 & 0.256 & 0.387 & 0.544 & - & - & - & - \\
      M-DCN \cite{zhang2020multi} & 0.345 & 0.255 & 0.380 & 0.528 & 0.475 & 0.440 & 0.485 & 0.540 \\
      GIE \cite{cao2022geometry} & \underline{0.362} & \underline{0.271} & \underline{0.401} & \underline{0.552} & \underline{0.491} & \textbf{0.452} & \underline{0.505} & 0.575 \\
      ReflectE \cite{zhang2022knowledge} & 0.358 & 0.263 & 0.396 & 0.546 & 0.488 & 0.450 & 0.501 & 0.559 \\
      \hline
      CompoundE & \textbf{0.367} & \textbf{0.275} & \textbf{0.402} & \textbf{0.555} & \textbf{0.493} & \underline{0.451} & \textbf{0.507} & \underline{0.578} \\
      \hline
    \end{tabular}
  \end{center}
\end{table*}

\section{Experiments}\label{sec:experiments}

{\bf Datasets.} We conduct experiments on three widely used benchmarking
datasets: ogbl-wikikg2, FB15k-237, and WN18RR. ogbl-wikikg2 is a
challenging Open Graph Benchmark dataset \cite{hu2020ogb} extracted from
the Wikidata \cite{vrandevcic2014wikidata} KG. Its challenge is to
design embedding models that can scale to large KGs. FB15k-237 and
WN18RR are extracted from the Freebase \cite{bollacker2008freebase} and
the WordNet \cite{miller1995wordnet}, respectively. Inverse relations
are removed from both to avoid the data leakage problem. Their main
challenge lies in modeling symmetry/antisymmetry and composition
relation patterns. The detailed statistics of the three datasets are
shown in Table \ref{tab:dataset_statistics} of the appendix. 

{\bf Evaluation Protocol.} To evaluate the link prediction performance
of CompoundE, we compute the rank of the ground truth entity in the list
of top candidates. Since embedding models tend to rank entities observed
in the training set higher, we compute the filtered rank to prioritize
candidates that would result in unseen triples. We follow the convention
and adopt the Mean Reciprocal Rank (MRR) and Hits@$k$ metrics to compare
the quality of different KGE models. Higher MRR and H@$k$ values
indicate better model performance. 

{\bf Performance Benchmarking.} Tables \ref{tab:wikikg2_overall} and
\ref{tab:fb15k-237_wn18rr_overall} show the best performance of
CompoundE and other benchmarking models for ogbl-wikikg2 and
FB15k-237/WN18RR datasets, respectively. The best results are shown in
bold fonts.  CompoundE consistently outperforms all benchmarking models
across all three datasets.  As shown in Table \ref{tab:wikikg2_overall},
the results of CompoundE are much better than previous KGE models while
the embedding dimension and the model parameter numbers are
significantly lower for the ogbl-wikikg2 dataset.  This implies lower
computation and memory costs of CompoundE.  We see from Table
\ref{tab:fb15k-237_wn18rr_overall} that CompoundE has achieved
significant improvement over distance-based KGE models using a single
operation, either translation (TransE), rotation (RotatE), or scaling
(PairRE). This confirms that cascading geometric transformations is an
effective strategy for designing KG embeddings. 

\begin{figure*}[ht!]
    \centering
    \subfloat[ogbl-wikikg2]{\includegraphics[width=0.33\textwidth]{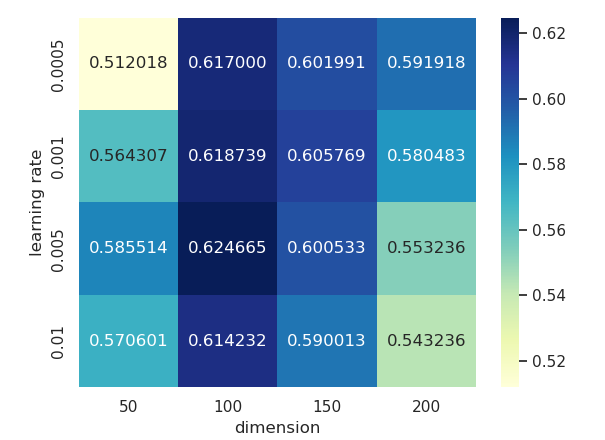}}
    \subfloat[FB15k-237]{\includegraphics[width=0.33\textwidth]{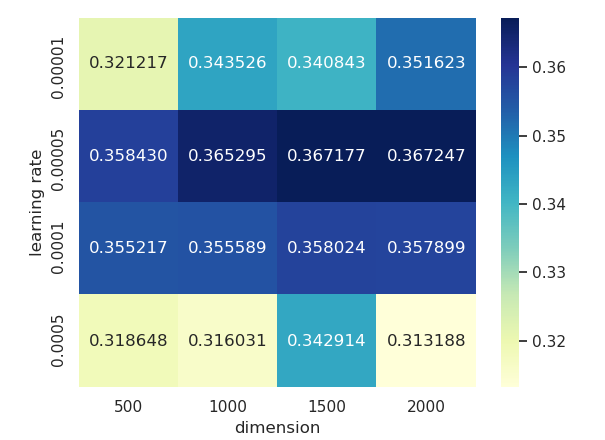}}
    \subfloat[WN18RR]{\includegraphics[width=0.33\textwidth]{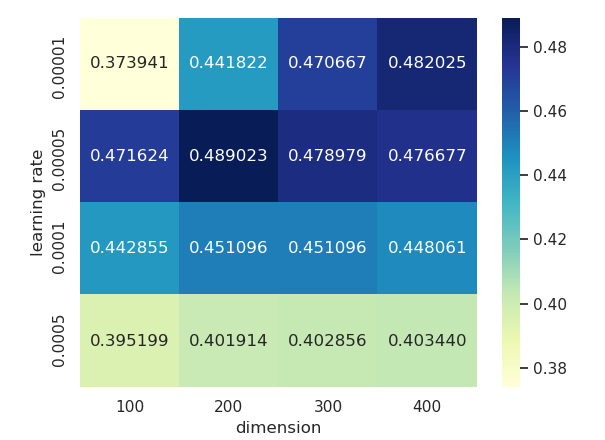}} 
    \caption{Heatmap of test MRR scores obtained from learning rate and dimension grid search.}
    \label{fig:wikikg_hyper_search}
\end{figure*}

{\bf Hyperparameter Tuning.} We conduct two sets of controlled
experiments to find the best model configurations for FB15k-237, WN18RR,
and ogbl-wikikg2 datasets. For the first set, we evaluate the effect of
different combinations of learning rates and embedding dimensions while
keeping the batch size, the negative sample size, and other parameters
constant. For the second set, we evaluate the effect of different
combinations of the training batch size and the negative sample size,
while keeping the learning rate, the embedding dimension, and other
parameters constant. Figs.  \ref{fig:wikikg_hyper_search} (a)-(c) show
MRR scores of CompoundE under different learning rates and embedding
dimension settings, for ogbl-wikikg2, FB15k-237, and WN18RR,
respectively. We see that CompoundE does not require a higher embedding
dimension to achieve the optimal performance, indicating that its model
size can be small. This is attractive for training and inference on
mobile/edge devices with limited memory capacity. The optimal model
configurations for three datasets are given in Table \ref{tab:optimal_configuration} of the appendix. 

{\bf Ablation Studies on CompoundE Variants.} We conduct ablation
studies on three variants of CompoundE; namely, CompoundE-Full,
CompoundE-Head and CompoundE-Tail as described in Section
\ref{subsec:definition}. The goal is to determine the variant that
performs the best on FB15k-237 and ogbl-wikikg2 datasets.  Moreover, we
test different ways of composing CompoundE by shuffling the order of
translation, scaling, and rotation operations.  Since geometric
transformations are not commutative, different orders of cascading yield
different models. Hence, we expect results for different CompoundE
variants to be different. We present results of distinct CompoundE
variants on ogbl-wikikg2 and FB15k-237 in Table \ref{tab:wikikg2_fb15k-237_variants} of the appendix. The main results
are summarized here. For ogbl-wikikg2, the best performing scoring
function is $\|\mathbf{h-\hat{S}\cdot \hat{T}\cdot\hat{R}\cdot t }\|$
while the best performing scoring function for FB15k-237 is
$\|\mathbf{S\cdot R\cdot T\cdot h - \hat{S}\cdot \hat{R}\cdot
\hat{T}\cdot t }\|$. For simplicity, we set the hyperparameters to be
the same across experiments in all variants. 

{\bf Performance on Complex Relation Types.} To gain insights into the
superior performance of CompoundE, we examine the performance of
CompoundE on each type of relations. KG relations can be categorized
into 4 types: 1) 1-to-1, 2) 1-to-N, 3) N-to-1, and 4) N-to-N. We can
classify relations by counting the co-occurrence of their respective
head and tail entities. A relation is classified as a 1-to-1 if each
head entity can co-occur with at most one tail entity, 1-to-N if each
head entity can co-occur with multiple tail entities, N-to-1 if multiple
head entities can co-occur with the same tail entity, and N-to-N if
multiple head entities can co-occur with multiple tail entities.  We
make decision based on the following rule. For each relation, $r$, we
compute the average number of subject (head) entities per object (tail)
entity as $hpt_r$ and the average number of object entities per subject
as $tph_r$. Then, with a specific threshold $\eta$,
\begin{equation}
    \begin{cases}
      hpt_r < \eta\; \text{and}\; tph_r< \eta \implies \text{$r$ is 1-to-1}\\
      hpt_r < \eta\; \text{and}\; tph_r\geq \eta \implies \text{$r$ is 1-to-N}\\
      hpt_r \geq \eta\; \text{and}\; tph_r< \eta \implies \text{$r$ is N-to-1}\\
      hpt_r \geq \eta\; \text{and}\; tph_r\geq \eta \implies \text{$r$ is N-to-N}.\\
    \end{cases}       
\end{equation}
We set $\eta=1.5$ as a logical threshold by following the convention.
Table \ref{tab:fb15k237_detail} compares the MRR scores of CompoundE
with benchmarking models on 1-to-1, 1-to-N, N-to-1, and N-to-N relations
in head and tail entities prediction performance for the FB15k237
dataset. We see that CompoundE consistently outperforms benchmarking
models in all relation categories. We show the performance of CompoundE
on each relation type for the WN18RR dataset in Table \ref{tab:wn18rr_detail} of the appendix.
Generally, CompoundE has a significant advantage over benchmarking
models for certain 1-N relations (e.g., ``member\_of\_domain\_usage''
and ``member\_of\_domain\_region'') and for some N-1 relations (e.g.,
``synset\_domain\_topic\_of'').  CompoundE is more effective than
traditional KGE models in modeling complex relations. 

\begin{figure}[ht!]
\centering
\includegraphics[width=0.6\columnwidth]{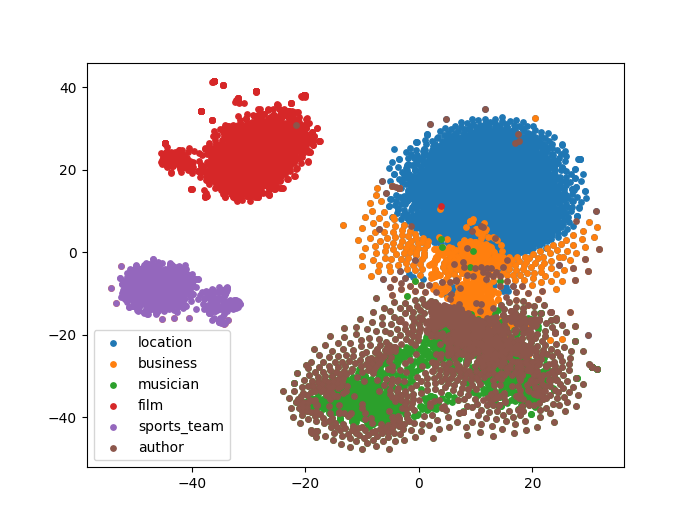}
\caption{$t$-SNE visualization of entity embedding in the 2D space for some
major entity types in FB15K-237.} \label{fig:entity_type}
\end{figure}

 \begin{table*}[ht!]
 \caption{Filtered MRR on four relation types of FB15k-237.}
  \begin{center}
    \label{tab:fb15k237_detail}
    \begin{tabular}{c|cccc|cccc} 
      \hline
      \textbf{Task} & \multicolumn{4}{c|}{\textbf{Predicting Head}} & \multicolumn{4}{c}{\textbf{Predicting Tail}}\\
      \hline
      \textbf{Rel. Category} & \textbf{1-to-1} & \textbf{1-to-N} & \textbf{N-to-1} & \textbf{N-to-N} & \textbf{1-to-1} & \textbf{1-to-N} & \textbf{N-to-1} & \textbf{N-to-N}\\
      \hline
      TransE & 0.492 & 0.454 & 0.081 & 0.252 & 0.485 & 0.072 & 0.740 & 0.367\\
      RotatE & 0.493 & 0.471 & 0.088 & 0.259 & 0.491 & 0.072 & 0.748 & 0.370\\
      PairRE & 0.496 & 0.476 & 0.117 & 0.274 & 0.492 & 0.073 & 0.763 & 0.387\\
      CompoundE & \textbf{0.501} & \textbf{0.488} & \textbf{0.123} & \textbf{0.279} & \textbf{0.497} & \textbf{0.074} & \textbf{0.783} & \textbf{0.394} \\
      \hline
    \end{tabular}
  \end{center}
\end{table*}

\begin{table*}[ht]
    \caption{Complexity comparison of KGE models.}
    \label{tab:complexity}
    \centering
    \begin{adjustbox}{width=0.95\textwidth,center}
    \begin{tabular}{cccccc} \hline
        Model & Ent. emb. & Rel. emb. & Scoring Function & Space & \# Params \\ \hline
TransE & $\mathbf{h}, \mathbf{t}\in\mathbb{R}^d$ & $\mathbf{r}\in\mathbb{R}^d$ 
& $-\left\|\mathbf{h}+\mathbf{r}-\mathbf{t}\right\|_{1/2}$ & $O((m+n)d)$ & 1251M \\
ComplEx & $\mathbf{h}, \mathbf{t}\in\mathbb{C}^d$ & $\mathbf{r}\in\mathbb{C}^d$ 
& $\text{Re}\left(\sum_{k=1}^{K} \mathbf{r}_k \mathbf{h}_{k} \overline{\mathbf{t}}_{k}\right)$ 
& $O((m+n)d)$ & 1251M \\
RotatE & $\mathbf{h}, \mathbf{t}\in\mathbb{C}^d$ & $\mathbf{r}\in\mathbb{C}^d$ 
& $-\left\|\mathbf{ h \circ r - t }\right\|$ & $O((m+n)d)$ & 1250M \\
PairRE & $\mathbf{h}, \mathbf{t}\in\mathbb{R}^d$ & $\mathbf{r^H, r^T}\in\mathbb{R}^d$ 
& $-\|\mathbf{h\odot r^H - t\odot r^T}\|$ & $O((m+n)d)$ & 500M \\ \hline
CompoundE-Head & $\mathbf{h}, \mathbf{t}\in\mathbb{R}^d$ & $\mathbf{T}[:, d-1], 
diag(\mathbf{S})\in\mathbb{R}^d, \mathbf{\theta}\in\mathbb{R}^{d/2}$ 
& $-\left\|\mathbf{T\cdot R(\theta)\cdot S\cdot h-t}\right\|$ & $O((m+n)d)$ & 250.1M\\ \\
CompoundE-Tail & $\mathbf{h}, \mathbf{t}\in\mathbb{R}^d$ & $\mathbf{\hat{T}}[:, d-1], 
diag(\mathbf{\hat{S}})\in\mathbb{R}^d, \mathbf{\theta}\in\mathbb{R}^{d/2}$ 
& $-\left\|\mathbf{h-\hat{T}\cdot \hat{R}(\theta)\cdot\hat{S}\cdot t }\right\|$ & $O((m+n)d)$ 
& 250.1M\\ \\
\multirow{2}{*}{CompoundE-Full} & \multirow{2}{*}{$\mathbf{h}, \mathbf{t}\in\mathbb{R}^d$} 
& $\mathbf{T}[:, d-1], diag(\mathbf{S})\in\mathbb{R}^d, \mathbf{\theta}\in\mathbb{R}^{d/2}$  
& \multirow{2}{*}{$-\left\|\mathbf{T\cdot R(\theta)\cdot S\cdot h - \hat{T}\cdot 
\hat{R}(\theta)\cdot \hat{S}\cdot t }\right\|$} & \multirow{2}{*}{$O((m+n)d)$} & \multirow{2}{*}{250.3M}\\
&  & $\mathbf{\hat{T}}[:,d-1], diag(\mathbf{\hat{S}})\in\mathbb{R}^d$  &  &  & \\ \hline
    \end{tabular}
    \end{adjustbox}
\end{table*}

\begin{figure*}[ht!]
    \centering
    \subfloat[$\mathbf{T}$]{\includegraphics[width=0.33\textwidth]{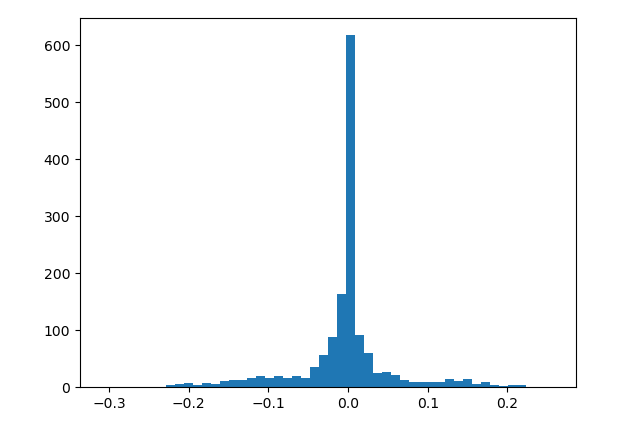}} 
    \subfloat[$\mathbf{R(\theta)}$]{\includegraphics[width=0.33\textwidth]{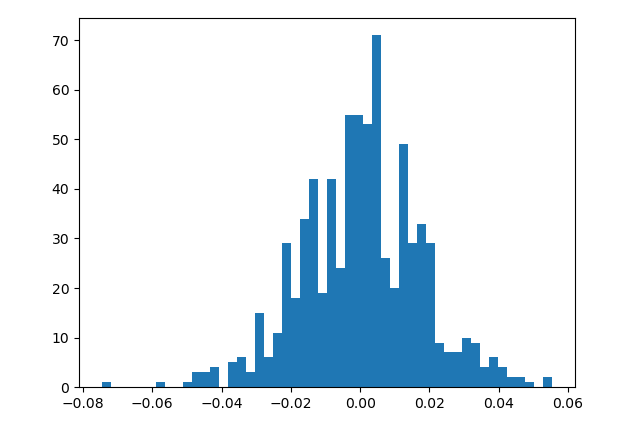}} 
    \subfloat[$\mathbf{S}$]{\includegraphics[width=0.33\textwidth]{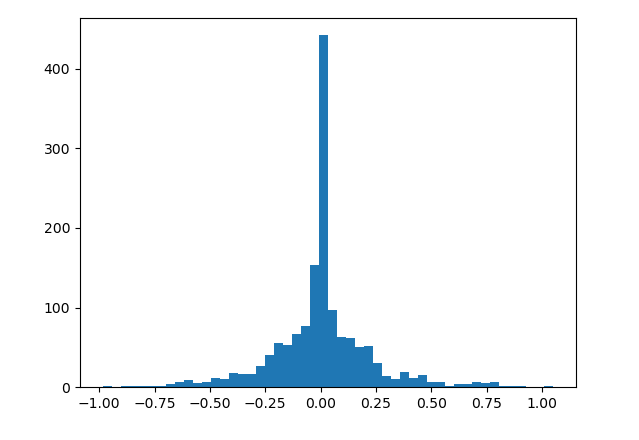}}
    \caption{Distribution of relation embedding values for ``Friends'' relation in FB15k-237, obtained using $\left\|\mathbf{S_r\cdot R_r\cdot T_r\cdot h-t}\right\|$}
    \label{fig:friend_oneside}
\end{figure*}

{\bf Complexity Analysis.} We compare the computational complexity of
CompoundE and several popular KGE models in Table \ref{tab:complexity}.
The last column gives the estimated number of free parameters used by
different models to achieve the best performance for the ogbl-wikikg2
dataset. CompoundE cuts the number of parameters at least by half while
achieving much better performance.  In the table, $n$, $m$, and $d$
denote the entity number, the relation number, and their embedding
dimension, respectively. Since $n\gg m$ in most datasets, we can afford
to increase the complexity of relation embedding for better link
prediction result without significantly increasing the overall model
complexity. 

{\bf Entity Semantics and Relation Component Values.} We provide a 2D
$t$-SNE visualization of the entity embedding generated by CompoundE for
FB15k-237 in Fig. \ref{fig:entity_type}.  Each entity is colored with
its respective entity type. As shown in the figure, some entity type
class are well separated while others are not. This scatter plot shows
that entity representations extracted by CompoundE can capture the
semantics of the entity. Thus, their embeddings can be used in various
downstream tasks such as KG entity typing and similarity based
recommendations.  In Fig. \ref{fig:friend_oneside}, we visualize
relation embedding for the ``friend'' relation in FB15k-237 by plotting
the histogram of translation, scaling, and rotation parameter values.
Since ``friend'' is a symmetric relation, we expect the translation
value to be close to zero, which is consistent with Fig.
\ref{fig:friend_oneside} (a).  Also, since ``friend'' is an N-to-N
relation, we expect the Compound operation to be singular. Actually,
most of the scaling values are zero as shown in Fig.
\ref{fig:friend_oneside} (c). They support our theoretical analysis of
CompoundE's properties. 

\section{Conclusion and Future Work}\label{sec:conclusion}

A new KGE model called CompoundE was proposed in this work.  We showed
that quite a few distance-based KGE models are special cases of
CompoundE. Extensive experiments were conducted for three datasets to
demonstrate the effectiveness of CompoundE. CompoundE achieves
state-of-the-art link prediction performance with a memory saving
solution for large KGs. Visualization of entity semantics and relation
embedding values was given to shed light on the superior performance of
CompoundE. 

We are interested in exploring two topics as future extensions.  First,
we may consider more complex operations in CompoundE. For example, there
is a recent trend to extend 2D rotations to 3D rotations for
rotation-based embeddings such as RotatE3D \cite{gao2020rotate3d}, SU2E
\cite{yang2020nage}. It is worthwhile to explore CompoundE-3D.  Second,
CompoundE is expected to be useful in many downstream tasks.  This
conjecture has to be verified. If this is the case, CompoundE can offer
a low memory solution to these tasks in realistic settings. 

\bibliography{abrv,conf_abrv,references}
\bibliographystyle{unsrt}  

\newpage

\section{Appendix}\label{sec:Appendix}
\subsection{Implementation and Optimal Configurations}

The statistics of the three datasets used in our experiments are
summarized in Table \ref{tab:dataset_statistics}. In the experiments, we
normalize all entity embeddings to unit vectors before applying compound
operations.  The optimal configurations of CompoundE are given in Table
\ref{tab:optimal_configuration}. The implementation of the rotation
operation in the optimal CompoundE configuration for the WN18RR dataset 
is adapted from RotatE. 

\begin{figure*}[ht!]
    \centering
    \subfloat[ogbl-wikikg2]{\includegraphics[width=0.33\textwidth]{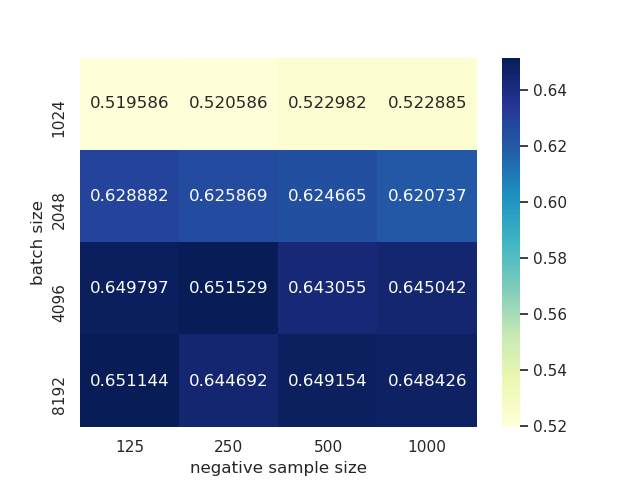}} 
    \subfloat[FB15k-237]{\includegraphics[width=0.33\textwidth]{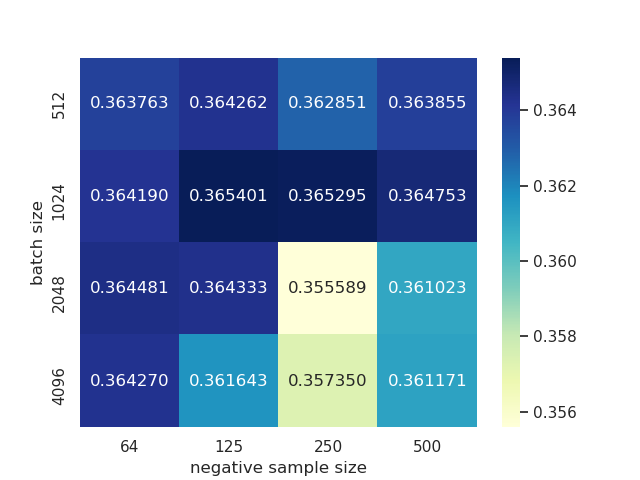}}
    \subfloat[WN18RR]{\includegraphics[width=0.33\textwidth]{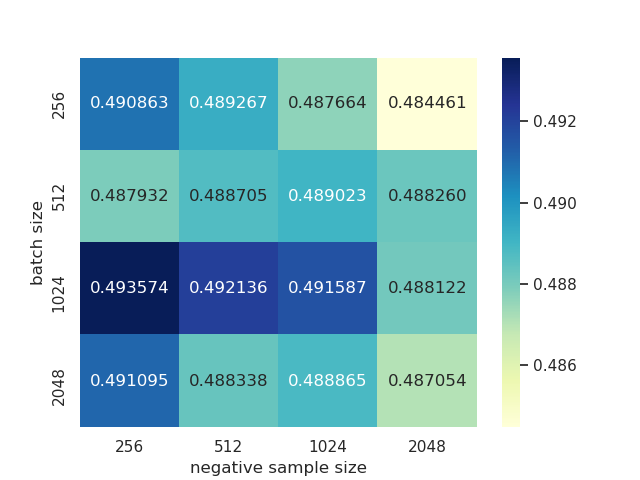}} 
    \caption{Heatmap of test MRR scores with a grid search for different combinations
of batch sizes and negative sample sizes.}\label{fig:wikikg_hyper_search_2}
\end{figure*}

More experimental results are given in Fig.
\ref{fig:wikikg_hyper_search_2}. The MRR scores of CompoundE under
different batch sizes and negative sampling sizes for ogbl-wikikg2,
FB15k-237, and WN18RR are shown in Figs.
\ref{fig:wikikg_hyper_search_2} (a)-(c), respectively.  All experiments
were conducted on a NVIDIA V100 GPU with 32GB memory. GPUs with larger
memory such as NVIDIA A100 (40GB), NVIDIA A40 (48GB) are only needed for
hyperparameter sweep when the dimension, the negative sample size, and
the batch size are high. We should point out that such settings are not
essential for CompoundE to obtain good results. They were used to search
for the optimal configurations. 

\begin{table*}[ht!]
    \caption{Datasets Statistics}\label{tab:dataset_statistics}
    \centering
    \begin{tabular}{cccccc}
    \hline
    \textbf{Dataset} & \textbf{\#Entities} & \textbf{\#Relations} & \textbf{\#Training} & \textbf{\#Validation} & \textbf{\#Test}\\
    \hline
    FB15k-237 & 14,541 & 237 & 272,115 & 17,535 & 20,466 \\
    WN18RR & 40,943 & 11 & 86,835 & 3,034 & 3,134 \\
    ogbl-wikikg2 & 2,500,604 & 535 & 16,109,182 & 429,456 & 598,543 \\
    \hline
    \end{tabular}
\end{table*}

\begin{table*}[ht]
    \caption{Optimal Configurations}\label{tab:optimal_configuration}
    \centering
    \begin{tabular}{cccccccc}
    \hline
    \textbf{Dataset} & \textbf{CompoundE Variant} & \textbf{\#Dim} & \textbf{\textit{lr}} & \textbf{\textit{B}} & \textbf{\textit{N}} & $\boldsymbol{\alpha}$ & $\boldsymbol{\zeta}$ \\
    \hline
    ogbl-wikikg2 & $\|\mathbf{h-\hat{S}\cdot \hat{T}\cdot\hat{R}\cdot t }\|$ & 100 & 0.005 & 4096 & 250 & 1 & 7\\
    FB15k-237 & $\|\mathbf{S\cdot R\cdot T\cdot h - \hat{S}\cdot \hat{R}\cdot \hat{T}\cdot t }\|$ & 1500 & 0.00005 & 1024 & 125 & 1 & 6\\
    WN18RR & $\|\mathbf{R\cdot S\cdot T\cdot h - \hat{S}\cdot \hat{T}\cdot t }\|$  & 200 & 0.00005 & 1024 & 256 & 0.5 & 6 \\
    \hline
    \end{tabular}
\end{table*}

\begin{figure*}[ht!]
\centering
\includegraphics[width=\textwidth]{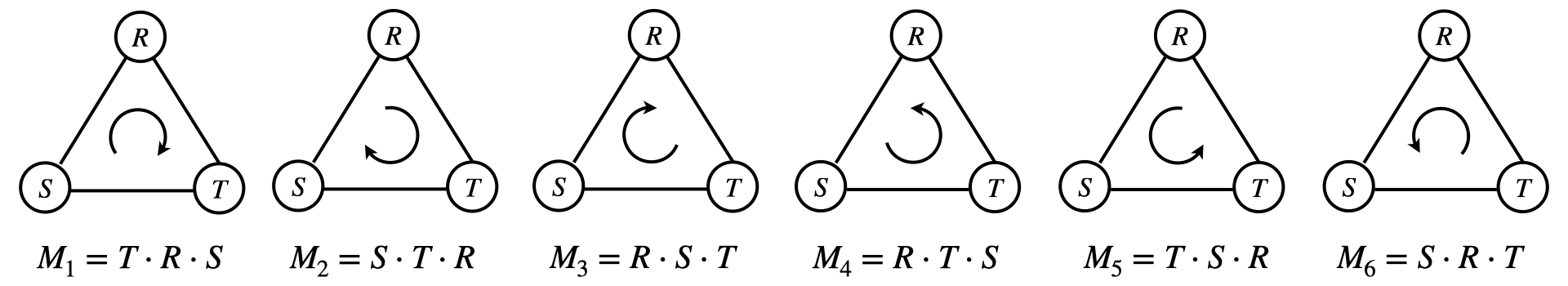}
\caption{Illustration of different ways of composing compound operations.}
\label{compoundE_permute}
\end{figure*}

\subsection{Compositions of CompoundE}

There are different ways to arrange the order of translation, rotation
and scaling operations in CompoundE. They are summarized in Fig.
\ref{compoundE_permute}. Furthermore, we report the link prediction
performance of different compositions of CompoundE for ogbl-wikikg2 and
FB15k-237 in Table \ref{tab:wikikg2_fb15k-237_variants}. The best
performing variant for ogbl-wikikg2 is $\|\mathbf{h-\hat{S}\cdot
\hat{T}\cdot\hat{R}\cdot t }\|$ and the second best is
$\|\mathbf{h-\hat{T}\cdot \hat{S}\cdot\hat{R}\cdot t }\|$. The most
effective variant for FB15k-237 $\|\mathbf{S\cdot R\cdot T\cdot h -
\hat{S}\cdot \hat{R}\cdot \hat{T}\cdot t }\|$ and the second best is
$\|\mathbf{T\cdot S\cdot R\cdot h - \hat{T}\cdot \hat{S}\cdot
\hat{R}\cdot t }\|$. 

\subsection{Properties of CompoundE}

Let $\mathbf{M}$ and $\mathbf{\hat{M}}$ denote the compound operation for
the head and tail entity embeddings, respectively. In the following, we will
prove nine properties of CompoundE.

\begin{prop}\label{prop1}
    CompoundE can model 1-N relations.
\end{prop}
\vspace{-2ex}
\begin{proof}
    A relation $r$ is an 1-N relation iff there exist at least two distinct tail entities $t_1$ and $t_2$ such that $(h, r, t_1)$ and $(h, r, t_2)$ both hold. Then we have:
    \begin{equation}
        \begin{aligned}
        \mathbf{M\cdot h} = \mathbf{\hat{M} \cdot t_1 }&,\quad \mathbf{M\cdot h} = \mathbf{\hat{M}\cdot t_2 }\\
        \mathbf{\hat{M}\cdot(t_1 - t_2)} &=  0
        \end{aligned}
    \end{equation}
    Since $\mathbf{t_1}\neq\mathbf{t_2}$, CompoundE can model 1-N relations when $\mathbf{\hat{M}}$ is singular.
\end{proof}

\begin{prop}\label{prop2}
    CompoundE can model N-1 relations.
\end{prop}
\vspace{-2ex}
\begin{proof}
    A relation $r$ is an N-1 relation iff there exist at least two distinct head entities $h_1$ and $h_2$ such that $(h_1, r, t)$ and $(h_1, r, t)$ both hold. Then we have:
    \begin{equation}
        \begin{aligned}
        \mathbf{M\cdot h_1} = \mathbf{\hat{M} \cdot t }&,\quad \mathbf{M\cdot h_2} = \mathbf{\hat{M}\cdot t }\\
        \mathbf{M\cdot(h_1-h_2)} &=  0
        \end{aligned}
    \end{equation}
    Since $\mathbf{h_1}\neq\mathbf{h_2}$, CompoundE can model N-1 relations when $\mathbf{M}$ is singular.
\end{proof}

\begin{prop}\label{prop3}
    CompoundE can model N-N relations.
\end{prop}
\vspace{-2ex}
\begin{proof}
    By the proof for Prop.\ref{prop1} and \ref{prop2}, N-N relations can be modeled when both $\mathbf{M}$ and $\mathbf{\hat{M}}$ are singular.
\end{proof}

\begin{prop}\label{prop4}
    CompoundE can model symmetric relations.
\end{prop}
\vspace{-2ex}
\begin{proof}
    A relation $r$ is a symmetric relation iff $(h, r, t)$ and $(t, r, h)$ holds simultaneously. Then we have:
    \begin{equation}
        \begin{aligned}
            \mathbf{M\cdot h} = \mathbf{\hat{M}\cdot t}&\implies\mathbf{h}=\mathbf{M^{-1}\hat{M}\cdot t} \\
            \mathbf{M\cdot t} = \mathbf{\hat{M}\cdot h}&\implies\mathbf{M\cdot t}=\mathbf{\hat{M}M^{-1}\hat{M}\cdot t}\\
            \mathbf{M}\mathbf{\hat{M}^{-1}} &= \mathbf{\hat{M}M^{-1}}
        \end{aligned}
    \end{equation}
    Therefore, CompoundE can model symmetric relations when $\mathbf{M}\mathbf{\hat{M}^{-1}} = \mathbf{\hat{M}M^{-1}}$.
\end{proof}

\begin{prop}\label{prop5}
    CompoundE can model antisymmetric relations.
\end{prop}
\vspace{-2ex}
\begin{proof}
    A relation $r$ is a antisymmetric relation iff $(h, r, t)$ holds but $(t, r, h)$ does not. By similar proof for Proposition \ref{prop4}, CompoundE can model symmetric relations when $\mathbf{M}\mathbf{\hat{M}^{-1}} \neq \mathbf{\hat{M}M^{-1}}$.
\end{proof}

\begin{prop}\label{prop6}
    CompoundE can model inversion relations.
\end{prop}
\vspace{-2ex}
\begin{proof}
    A relation $r_2$ is the inverse of relation $r_1$ iff $(h, r_1, t)$ and $(t, r_2, h)$ holds simultaneously. Then we have:
    \begin{equation}
        \begin{aligned}
            \mathbf{M_1\cdot h}=\mathbf{\hat{M}_1\cdot t}&\implies\mathbf{h}=\mathbf{M_1^{-1}\hat{M}_1\cdot t}\\
            \mathbf{M_2\cdot t =\hat{M}_2\cdot h}&\implies\mathbf{M_2\cdot t}=\mathbf{\hat{M}_2 M_1^{-1}\hat{M}_1\cdot t}\\
            \mathbf{\hat{M}_2^{-1}M_2} &= \mathbf{ M_1^{-1}\hat{M}_1}\\
        \end{aligned}
    \end{equation}
    Therefore, CompoundE can model inversion relations when $\mathbf{\hat{M}_2^{-1}M_2} = \mathbf{ M_1^{-1}\hat{M}_1}$.
\end{proof}

\begin{prop}\label{prop7}
  CompoundE can model transitive relations.
\end{prop}
\vspace{-2ex}
\begin{proof}
    A set of relations $(r_1, r_2, r_3)$ are transitive iff $(e_1, r_1, e_2)$,  $(e_2, r_2, e_3)$, and $(e_1, r_3, e_3)$ hold simultaneously. Then we have:
    \begin{equation}
    \begin{aligned}
        \mathbf{M_1 \cdot e_1 = \hat{M}_1 \cdot e_2} &\implies \mathbf{e_1 = M_1^{-1}\hat{M}_1 \cdot e_2}\\
        \mathbf{M_2 \cdot e_2 = \hat{M}_2 \cdot e_3} &\implies \mathbf{e_3 = \hat{M}_2^{-1} M_2 \cdot e_2}\\
        \mathbf{M_3 \cdot e_1} &= \mathbf{\hat{M}_3 \cdot e_3} \\
        \mathbf{M_3 M_1^{-1}\hat{M}_1 \cdot e_2} &= \mathbf{\hat{M}_3\hat{M}_2^{-1} M_2 \cdot e_2}\\
        \mathbf{\hat{M}_3^{-1}M_3} &= \mathbf{(\hat{M}_2^{-1} M_2)(\hat{M}_1^{-1}M_1)}
    \end{aligned}
    \end{equation}
    Therefore, CompoundE can model transitive relations when $\mathbf{\hat{M}_3^{-1}M_3} = \mathbf{(\hat{M}_2^{-1} M_2)(\hat{M}_1^{-1}M_1)}$.
\end{proof}

\begin{table*}[ht!]
 \caption{Filtered ranking of link prediction for ogbl-wikikg2 and FB15k-237}
  \begin{center}
    \label{tab:wikikg2_fb15k-237_variants}
    \begin{tabular}{c|cccc|cccc} 
      \hline
      \textbf{Datasets} & \multicolumn{4}{c|}{\textbf{ogbl-wikikg2}} & \multicolumn{4}{c}{\textbf{FB15k-237}} \\
      \hline
      \textbf{Model} & \textbf{MRR} & \textbf{Hit@1} & \textbf{Hit@3} & \textbf{Hit@10} & \textbf{MRR} & \textbf{Hit@1} & \textbf{Hit@3} & \textbf{Hit@10}\\
      \hline
      $\left\|\mathbf{T\cdot R\cdot S\cdot h-t}\right\|$ & 0.6001 & 0.5466 & 0.6187 & 0.7043 & 0.3373 & 0.2455 & 0.3720 & 0.5217 \\
      $\left\|\mathbf{T\cdot S\cdot R\cdot h-t}\right\|$ & 0.5972 & 0.5431 & 0.6157 & 0.7002 & 0.3359 & 0.2467 & 0.3685 & 0.5171 \\
      $\left\|\mathbf{S\cdot T\cdot R\cdot h-t}\right\|$ & 0.6019 & 0.5459 & 0.6211 & 0.7091 & 0.3354 & 0.2461 & 0.3680 & 0.5169 \\
      $\left\|\mathbf{R\cdot T\cdot S\cdot h-t}\right\|$ & 0.5838 & 0.5288 & 0.6016 & 0.6880 & 0.3356 & 0.2456 & 0.3687 & 0.5167 \\
      $\left\|\mathbf{S\cdot R\cdot T\cdot h-t}\right\|$ & 0.6006 & 0.5460 & 0.6185 & 0.7043 & 0.3342 & 0.2449 & 0.3662 & 0.5141 \\
      $\left\|\mathbf{R\cdot S\cdot T\cdot h-t}\right\|$ & 0.5834 & 0.5239 & 0.6039 & 0.6979 & 0.3355 & 0.2460 & 0.3698 & 0.5165 \\
      \hline
      $\|\mathbf{h-\hat{T}\cdot \hat{R}\cdot\hat{S}\cdot t }\|$ & 0.6440 & 0.5777 & 0.6688 & 0.7787 & 0.3326 & 0.2393 & 0.3681 & 0.5183 \\
      $\|\mathbf{h-\hat{T}\cdot \hat{S}\cdot\hat{R}\cdot t }\|$ & \underline{0.6497} & \underline{0.5827} & \underline{0.6758} & \underline{0.7858} & 0.3302 & 0.2384 & 0.3658 & 0.5123 \\
      $\|\mathbf{h-\hat{S}\cdot \hat{T}\cdot\hat{R}\cdot t }\|$ & \textbf{0.6515} & \textbf{0.5844} & \textbf{0.6781} & \textbf{0.7873} & 0.3313 & 0.2397 & 0.3680 & 0.5113 \\
      $\|\mathbf{h-\hat{R}\cdot \hat{T}\cdot\hat{S}\cdot t }\|$ & 0.6434 & 0.5779 & 0.6678 & 0.7763 & 0.3312 & 0.2394 & 0.3674 & 0.5136 \\
      $\|\mathbf{h-\hat{S}\cdot \hat{R}\cdot\hat{T}\cdot t }\|$ & 0.6474 & 0.5802 & 0.6739 & 0.7842 & 0.3290 & 0.2384 & 0.3640 & 0.5090 \\
      $\|\mathbf{h-\hat{R}\cdot \hat{S}\cdot\hat{T}\cdot t }\|$ & 0.6442 & 0.5777 & 0.6699 & 0.7788 & 0.3298 & 0.2383 & 0.3650 & 0.5114 \\
      \hline
      $\|\mathbf{T\cdot R\cdot S\cdot h - \hat{T}\cdot \hat{R}\cdot \hat{S}\cdot t }\|$ & 0.5479 & 0.4918 & 0.5661 & 0.6549 & 0.3426 & 0.2535 & 0.3770 & 0.5201 \\
      $\|\mathbf{T\cdot S\cdot R\cdot h - \hat{T}\cdot \hat{S}\cdot \hat{R}\cdot t }\|$ & 0.5776 & 0.5210 & 0.5975 & 0.6852 & \underline{0.3613} & \underline{0.2702} & \underline{0.3948} & \underline{0.5462} \\
      $\|\mathbf{S\cdot T\cdot R\cdot h - \hat{S}\cdot \hat{T}\cdot \hat{R}\cdot t }\|$ & 0.5782 & 0.5249 & 0.5962 & 0.6783 & 0.3597 & 0.2687 & 0.3937 & 0.5450 \\
      $\|\mathbf{R\cdot T\cdot S\cdot h - \hat{R}\cdot \hat{T}\cdot \hat{S}\cdot t }\|$ & 0.5611 & 0.5053 & 0.5802 & 0.6660 & 0.3402 & 0.2506 & 0.3721 & 0.5213 \\
      $\|\mathbf{S\cdot R\cdot T\cdot h - \hat{S}\cdot \hat{R}\cdot \hat{T}\cdot t }\|$ & 0.5736 & 0.5175 & 0.5918 & 0.6805 & \textbf{0.3634} & \textbf{0.2718} & \textbf{0.3984} & \textbf{0.5500} \\
      $\|\mathbf{R\cdot S\cdot T\cdot h - \hat{R}\cdot \hat{S}\cdot \hat{T}\cdot t }\|$ & 0.5586 & 0.5057 & 0.5743 & 0.6592 & 0.3493 & 0.2593 & 0.3836 & 0.5301 \\
      \hline
    \end{tabular}
  \end{center}
\end{table*}

  \begin{table*}[ht!]
  \caption{Filtered MRR on each relation type of WN18RR}\label{tab:wn18rr_detail}
  \begin{center}
    \begin{tabular}{l|cccc} 
      \hline
      \textbf{Relation} & \textbf{Category} & \textbf{TransE} & \textbf{RotatE} & \textbf{CompoundE}\\
      \hline
      similar\_to & 1-to-1 & 0.294 & \textbf{1.000} & \textbf{1.000}\\
      verb\_group & 1-to-1 & 0.363 & 0.961 & \textbf{0.974} \\
      member\_meronym & 1-to-N & 0.179 & \textbf{0.259} & 0.230 \\
      has\_part & 1-to-N & 0.117 & \textbf{0.200} & 0.190 \\
      member\_of\_domain\_usage & 1-to-N & 0.113 & 0.297 & \textbf{0.332} \\
      member\_of\_domain\_region & 1-to-N & 0.114 & 0.217 & \textbf{0.280} \\
      hypernym & N-to-1 & 0.059 & \textbf{0.156} & 0.155 \\
      instance\_hypernym & N-to-1 & 0.289 & 0.322 & \textbf{0.337} \\
      synset\_domain\_topic\_of & N-to-1 & 0.149 & 0.339 & \textbf{0.367} \\
      also\_see & N-to-N & 0.227 & 0.625 & \textbf{0.629} \\
      derivationally\_related\_form & N-to-N & 0.440 & \textbf{0.957} & 0.956 \\
      \hline
    \end{tabular}
  \end{center}
\end{table*}

\begin{prop}\label{prop8}
  CompoundE can model both both commutative and non-commutative relations. 
\end{prop}
\vspace{-2ex}
\begin{proof}
  Since the general form of affine group is non-commutative, our proposed CompoundE is non-commutative i.e.
  \begin{equation}
      (\mathbf{M_1\hat{M}_1^{-1}})(\mathbf{M_2\hat{M}_2^{-1}})\neq (\mathbf{M_2\hat{M}_2^{-1}})(\mathbf{M_1\hat{M}_1^{-1}})
  \end{equation}
  where each $\mathbf{M}$ consists of translation, rotation, and scaling component. 
  However, in special cases, when our relation embedding has only one of the translation, rotation, or scaling component, then the relation embedding becomes commutative again.
\end{proof}

\begin{prop}\label{prop9}
  CompoundE can model sub-relations.
\end{prop}
\vspace{-2ex}
\begin{proof}
A relation $r_1$ is a sub-relation of $r_2$ if $(h, r_2, t)$ implies $(h, r_1, t)$. Without loss of generality, suppose our compounding operation takes the following form
\begin{equation}
    \mathbf{M = T\cdot R\cdot S, \; \hat{M} = \hat{T}\cdot \hat{R}\cdot \hat{S},} 
\end{equation}
and suppose
\begin{equation}
    \begin{aligned}
        &\mathbf{T_1 = T_2, \hat{T}_1 = \hat{T}_2,} \\
        &\mathbf{R_1 = R_2, \hat{R}_1 = \hat{R}_2,}\\
        &\mathbf{S_1 = \gamma S_2, \hat{S}_1 = \gamma \hat{S}_2}, \gamma\leq 1.\\
    \end{aligned}
\end{equation}
With these conditions, we can compare the CompoundE 
scores generated by $(h, r_1, t)$ and  $(h, r_2, t)$ as follows:
\begin{equation}
    \begin{aligned}
        &f_{r_1}(h,t) - f_{r_2}(h,t) \\
        =& \|\mathbf{T_1\cdot R_1\cdot S_1\cdot h - \hat{T}_1\cdot \hat{R}_1\cdot \hat{S}_1\cdot t }\|- \|\mathbf{T_2\cdot R_2\cdot S_2\cdot h - \hat{T}_2\cdot \hat{R}_2\cdot \hat{S}_2\cdot t }\| \\
        =& \|\mathbf{T_2\cdot R_2 \cdot (\gamma S_2) \cdot h - \hat{T}_2\cdot \hat{R}_2 \cdot (\gamma \hat{S}_2)\cdot t }\|- \|\mathbf{T_2\cdot R_2\cdot S_2\cdot h - \hat{T}_2\cdot \hat{R}_2\cdot \hat{S}_2\cdot t }\| \\
        =& \|\gamma(\mathbf{T_2\cdot R_2\cdot S_2\cdot h - \hat{T}_2\cdot \hat{R}_2\cdot \hat{S}_2\cdot t })\|- \|\mathbf{T_2\cdot R_2\cdot S_2\cdot h - \hat{T}_2\cdot \hat{R}_2\cdot \hat{S}_2\cdot t }\| \leq 0
    \end{aligned}
\end{equation}
This means that $(h, r_1, t)$ generates a smaller error score than $(h, r_2, t)$. If $(h, r_2, t)$ holds, $(h, r_1, t)$ must also holds. Therefore, $r_1$ is a sub-relation of $r_2$.
\end{proof} 
    
\begin{figure}[ht!]
    \centering
    \subfloat[]{\includegraphics[width=0.2\textwidth]{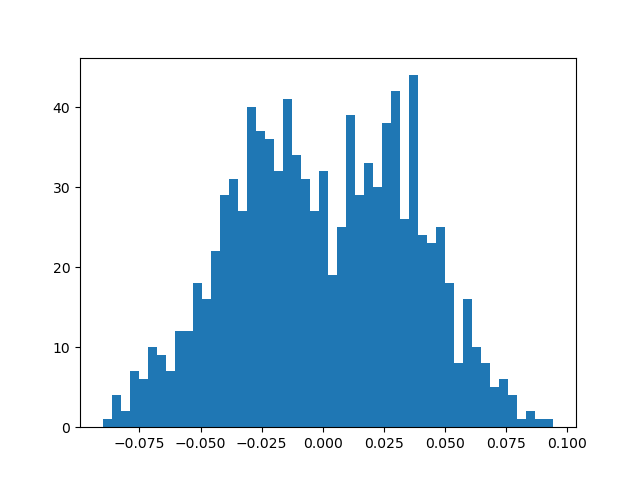}}
    \subfloat[]{\includegraphics[width=0.2\textwidth]{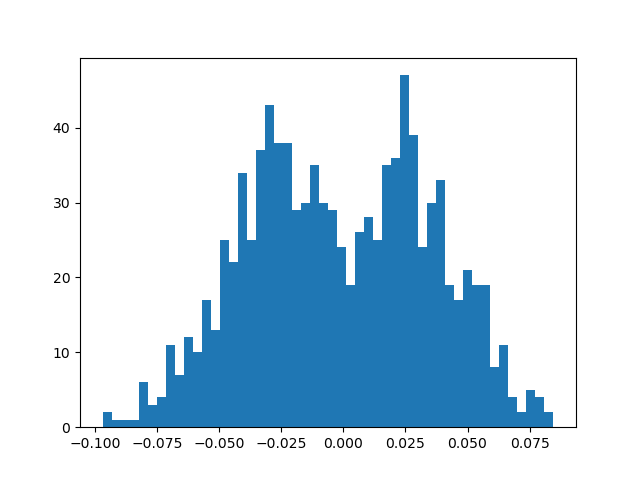}}
    \subfloat[]{\includegraphics[width=0.2\textwidth]{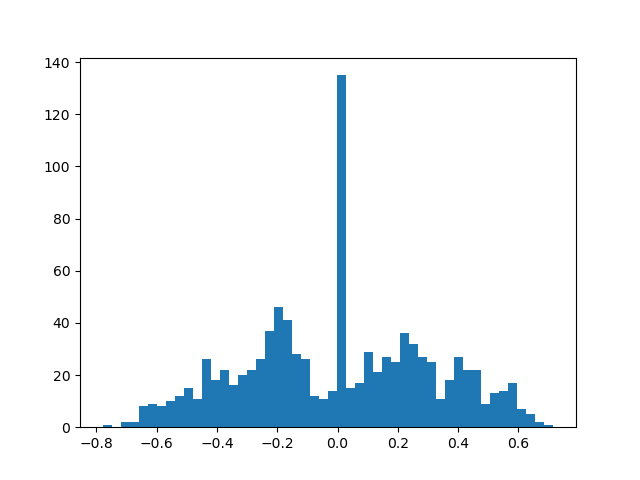}}
    \subfloat[]{\includegraphics[width=0.2\textwidth]{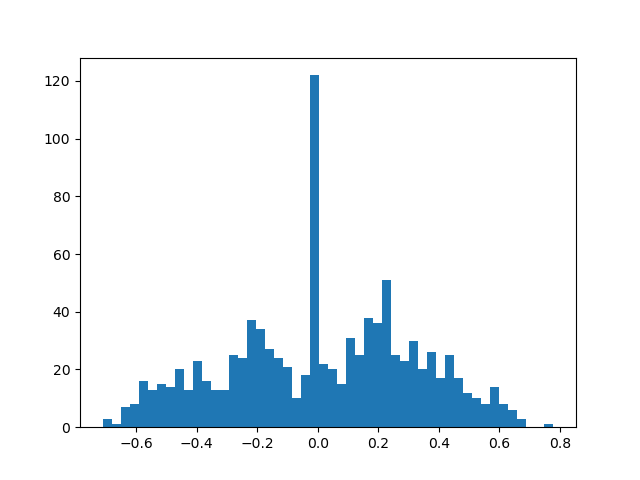}}
    \subfloat[]{\includegraphics[width=0.2\textwidth]{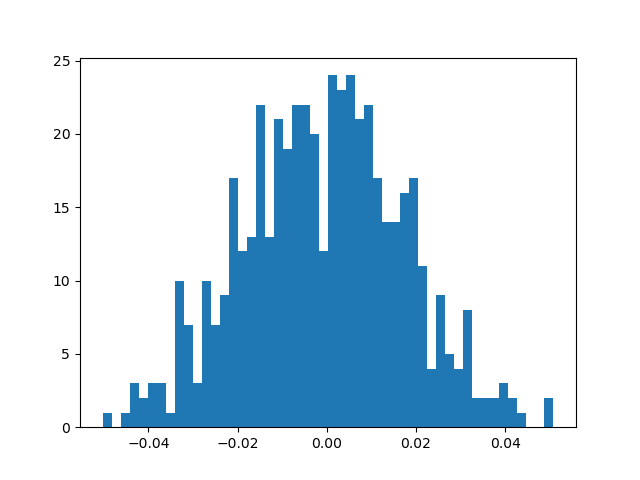}} 
\caption{FB15k-237 ``Friends'' relation embedding obtained using
$\|\mathbf{S\cdot R\cdot T\cdot h - \hat{S}\cdot \hat{R}\cdot
\hat{T}\cdot t }\|$: (a) distribution of head translation values, (b)
distribution of tail translation values, (c) distribution of head scaling
values, (d) distribution of tail scaling values, and (e) distribution of
rotation angle values.} \label{fig:friend_full}
\end{figure}

\begin{figure}[ht!]
    \centering
    \subfloat[]{\includegraphics[width=0.25\textwidth]{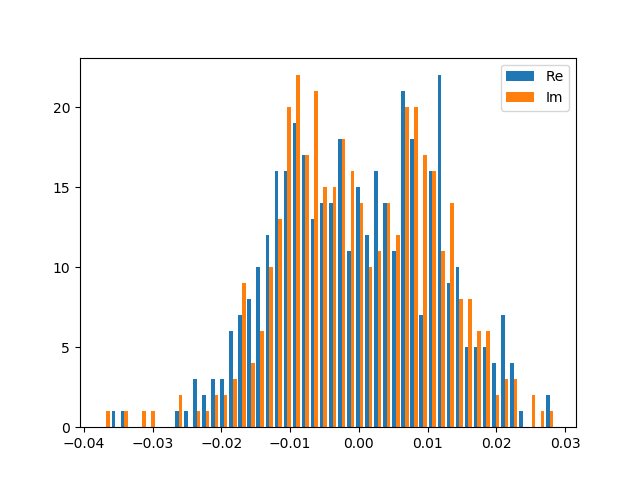}}
    \subfloat[]{\includegraphics[width=0.25\textwidth]{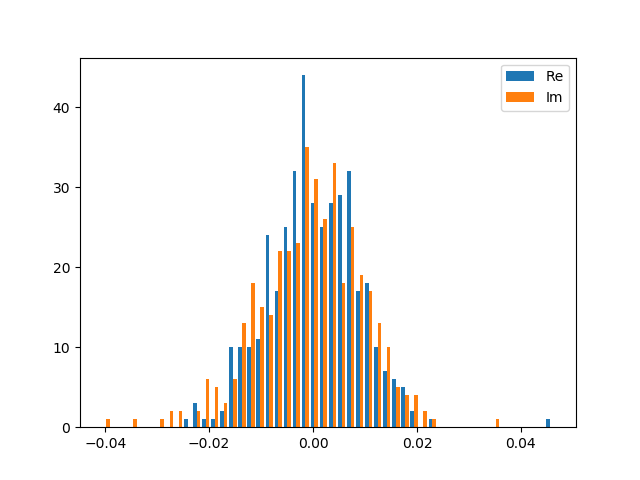}}
    \subfloat[]{\includegraphics[width=0.25\textwidth]{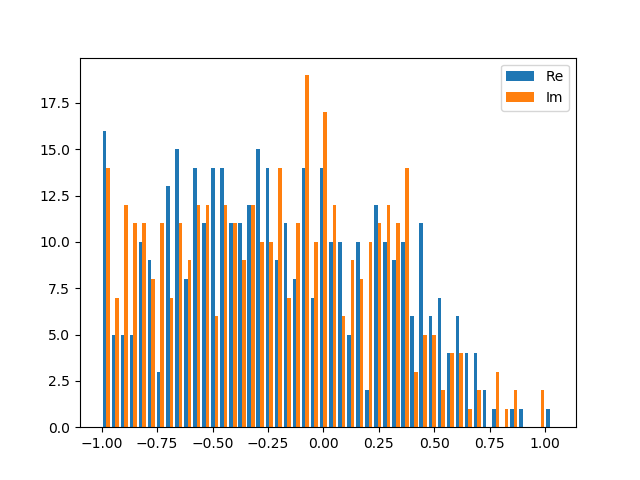}}
    \subfloat[]{\includegraphics[width=0.25\textwidth]{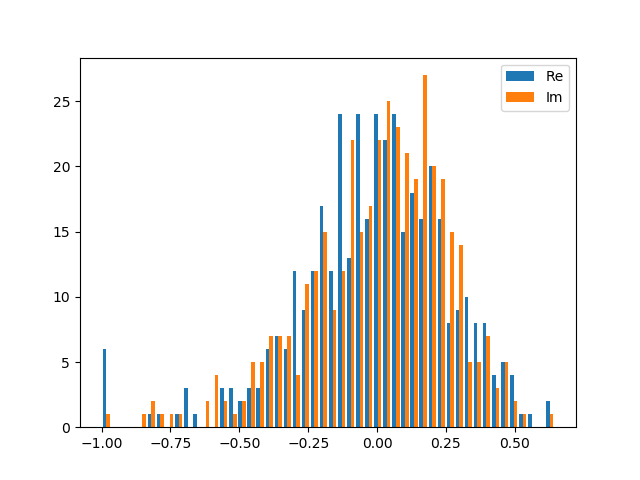}}
\caption{WN18RR ``instance\_hypernym'' relation: (a) distribution of head
translation values, (b) distribution of tail translation values, (c)
distribution of head scaling values, and (d) distribution of tail scaling
values.} \label{fig:wn18rr_antisymmetric}
\end{figure}

\begin{figure}[ht!]
    \centering
    \subfloat[]{\includegraphics[width=0.25\textwidth]{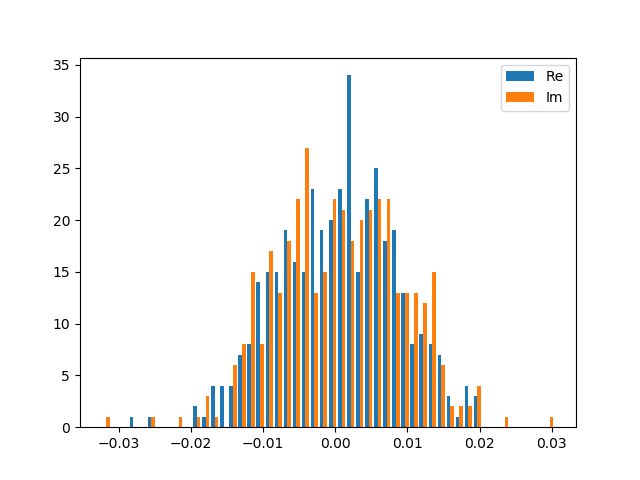}}
    \subfloat[]{\includegraphics[width=0.25\textwidth]{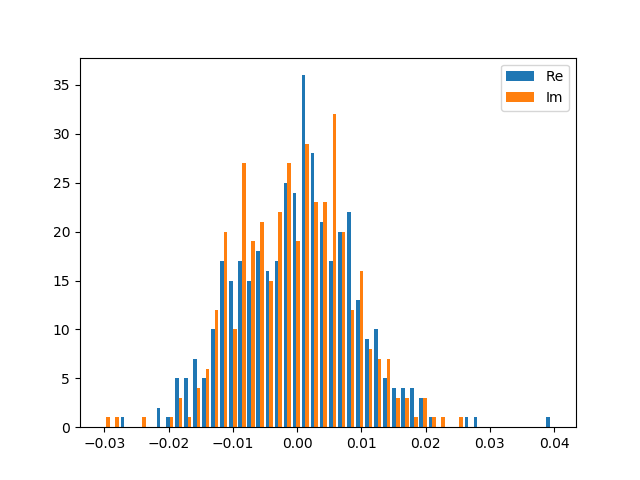}}
    \subfloat[]{\includegraphics[width=0.25\textwidth]{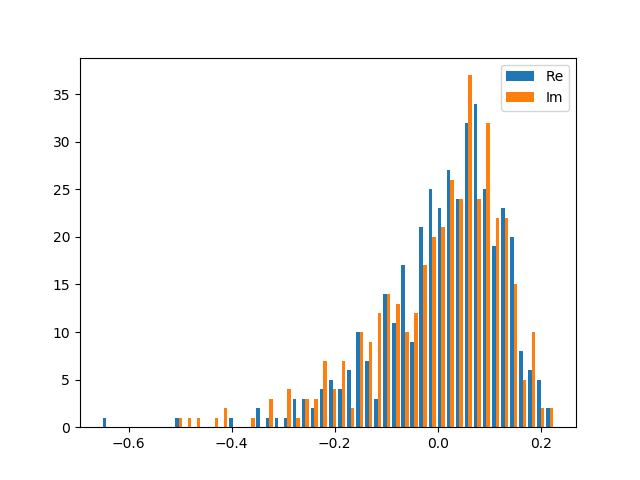}}
    \subfloat[]{\includegraphics[width=0.25\textwidth]{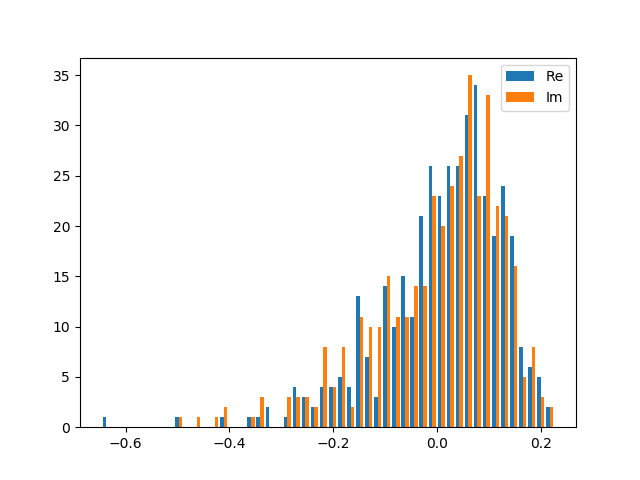}}
\caption{WN18RR ``similar\_to'' relation: (a) distribution of head
translation values, (b) distribution of tail translation values, (c)
distribution of head scaling values, and (d) distribution of tail scaling
values.}\label{fig:wn18rr_symmetric}
\end{figure}

\subsection{Complex Relation Modeling and Histograms of Embedding Values}

The filtered MRR scores on each relation type of WN18RR are given in
Table \ref{tab:wn18rr_detail}. We see that CompoundE has a significant
advantage over benchmarking models for certain 1-N relations such as
``member\_of\_domain\_usage'' ($+11.8\%$) and
``member\_of\_domain\_region'' ($+29.0\%$) and for some N-1 relations
such as ``synset\_domain\_topic\_of'' ($+8.3\%$). 

Besides the histograms shown in the main paper, we add more plots to
visualize CompoundE relation embedding values.  In Fig.
\ref{fig:friend_full}, we show the embedding values for the ``Friends''
relation in the FB15k-237. We use the CompoundE-full variant
($\|\mathbf{S_r\cdot R_r\cdot T_r\cdot h-\hat{S}_r\cdot \hat{R}_r\cdot
\hat{T}_r\cdot t}\|$) to generate the embedding. We plot the translation
and scaling components for both the head and the tail.  We only show a
single plot for the rotation component since the rotation parameter is
shared between the head and the tail. Different from the CompoundE-head
($\|\mathbf{S_r\cdot R_r\cdot T_r\cdot h-t}\|$), we see two modes
(instead of only one mode) in CompoundE-full's plots. One conjecture for
this difference is that CompoundE-full has a pair of operations on both
the head and the tail, the distribution of values need to have two modes
to maintain the symmetry. Similar to CompoundE-head, the scaling
parameters of CompoundE-full have a large amount of zeros to maintain
the singularity of compounding operators and help learn the N-to-N
complex relations. 

Fig. \ref{fig:wn18rr_antisymmetric} and Fig. \ref{fig:wn18rr_symmetric}
display the histogram of relation embeddings for ``instance\_hypernym''
relation and ``similar\_to'' relation in WN18RR, respectively.  The real
(in blue) and the imaginary (in orange) parts are overlaid in each plot.
Notice that ``instance\_hypernym'' is an antisymmetric relation while
``similar\_to'' is a symmetric relation. This relation pattern is
reflected on the embedding histogram since the translation and the
scaling histograms for the head and the tail are different in
``instance\_hypernym''.  In contrast, the translation and scaling
histograms for the head and the tail are almost identical in
``similar\_to''. 

\end{document}